\documentclass[11pt]{article}
\usepackage{acl2016}
\usepackage{flushend}
\usepackage{times}
\usepackage{latexsym}
\usepackage{graphicx}
\usepackage{color}
\usepackage{algorithm2e}
\usepackage{tabularx}
\usepackage{amsfonts}

\usepackage{framed,multirow}

\usepackage[utf8]{inputenc}
\usepackage{CJKutf8}

\newenvironment{SChinese}{%                                                      
  \CJKfamily{gbsn}%                                                             
  \CJKtilde
  \CJKnospace}{}

\aclfinalcopy % Uncomment this line for the final submission
\title{Morphology Generation for Statistical Machine Translation\\ using Deep Learning Techniques}
 
\author{Marta R. Costa-juss\`a and Carlos Escolano \\
         TALP Research Center \\ Universitat Polit\`ecnica de Catalunya, Barcelona \\
  {\tt marta.ruiz@upc.edu, carlos.escolano@tsc.upc.edu}}

\date{}

\begin{document}

\maketitle

\begin{abstract}
Morphology in unbalanced languages remains a big challenge in the context of machine translation. In this paper, we propose to de-couple machine translation from morphology generation in order to better deal with the problem. We investigate the morphology simplification with a reasonable trade-off between expected gain and generation complexity. For the Chinese-Spanish task, optimum morphological simplification is in gender and number. For this purpose, we design a new classification architecture which, compared to other standard machine learning techniques, obtains the best results. This proposed neural-based architecture consists of several layers: an embedding, a convolutional followed by a recurrent neural network and, finally, ends with sigmoid and softmax layers. We obtain classification results over 98\% accuracy in gender classification, over 93\% in number classification, and an overall translation improvement of 0.7 METEOR. 
\end{abstract}

\section{Introduction}
%MARTA

\label{sec:introduction}

Machine Translation (MT) is evolving from different perspectives. One of the most popular paradigms is still Statistical Machine Translation (SMT), which consists in finding the most probable target sentence given the source sentence using probabilistic models based on co-ocurrences. Recently, deep learning techniques applied to natural language processing, speech recognition and image processing and even in MT have reached quite successful results. Early stages of deep learning applied to MT include using neural language modeling for rescoring \cite{schwenk:2007}. Later, deep learning has been integrated in MT in many different steps \cite{survey}. Nowadays, deep learning has allowed to develop an entire new paradigm, which within one-year of development has achieved state-of-the-art results \cite{jean:2015} for some language pairs.

In this paper, we are focusing on a challenging translation task, which is Chinese-to-Spanish. This translation task has the characteristic that we are going from an isolated language in terms of morphology (Chinese) to a fusional language (Spanish). This means that for a simple word in Chinese (e.g. \textit{\begin{CJK}{UTF8}{} \begin{SChinese}鼓励  \end{SChinese} \end{CJK}}), the corresponding translation has many different morphology inflexions (e.g. \textit{alentar, alienta, alentamos, alientan ...}), which depend on the context. It is still difficult for MT in general (no matter which paradigm) to extract information from the source context to give the correct translation.

We propose to divide the problem of translation into translation and then a postprocessing of morphology generation. This has been done before, e.g. \cite{toutanova:2008,formiga:2013}, as we will review in the next section. However, the main contribution of our work is that we are using deep learning techniques in morphology generation. This gives us significant improvements in translation quality.

The rest of the paper is organised as follows. Section \ref{sec:relwork} describes the related work both in morphology generation approaches and in Chinese-Spanish translation. Section \ref{sec:mt} overviews the phrase-based MT approach together with an explanation of the divide and conquer approach of translating and generating morphology. Section \ref{sec:mg} details the architecture of the morphology generation module and it reports the main classification techniques that are used for morphology generation. Section \ref{sec:experimental} describes the experimental framework. Section \ref{sec:evaluation} reports and discusses both classification and translation results, which show significant improvements. Finally, section \ref{sec:conclusions} summarises the main conclusions and further work.

\section{Related Work}
%MARTA

\label{sec:relwork}

In this section we are reviewing the previous related works on morphology generation for MT and on Chinese-Spanish MT approaches.

\paragraph{Morphology generation} There have been many works in morphological generation and some of them are in the context of the application of MT. In this cases, MT is faced in two-steps: first step where the source is translated to a simplified target text that has less morphology variation than the original target; and then, second step, a postprocessing module (morphology generator) adds the proper inflections. 
To name a few of these works, for example, \cite{toutanova:2008} build maximum entropy markov models for inflection prediction of stems;  \cite{clifton:2011} and \cite{kholy:2012} use conditional random fields (CFR) to predict one or more morphological features; and \cite{formiga:2013} use Support Vector Machines (SVMs) to predict verb inflections.
Other related works are in the context of Part-of-Speech (PoS) tagging generation such as \cite{gimenez2004svmtool} in which a model is trained to predict each individual fragment of a PoS tag by means of machine learning algorithms. The main difference is that in PoS tagging the word itself has information about morphological inflection, whereas in our task, we do not have this information.

In this paper, we use deep learning techniques to morphology generation or classification. Based on the fact that Chinese does not have number and gender inflections and Spanish does, \cite{costajussa:2015} show that simplification in gender and number has the best trade-off between improving translation and keeping the morphology generation complexity at a low level.

\paragraph{Chinese-Spanish} There are few works in Chinese-Spanish MT despite being two of the most spoken languages in the world. Most of these works are based on comparing different pivot strategies like standard cascade or pseudo-corpus \cite{Marta2012}. Also it is important to mention that, in 2008, there were two tasks organised by the popular IWSLT evaluation campaign\footnote{http://iwslt2010.fbk.eu} (International Workshop on Spoken Language Translation) between these two languages \cite{iwslt}. The first task was based on a direct translation for Chinese-Spanish. The second task provided corpus in Chinese-English and English-Spanish and asked participants to provide Chinese-Spanish translation through pivot techniques. The second task obtained better results than direct translation because of the larger corpus provided. Differently, \cite{Jordi2016} present the first rule-based MT system for Chinese to Spanish. Authors describe a hybrid method for constructing this system taking advantage of available resources such as parallel corpora that are used to extract dictionaries and lexical and structural transfer rules. Finally, it is worth mentioning that novel successful neural approximations \cite{jean:2015}, already mentioned in the introduction, have not yet achieved state-of-the-art results for this language pair \cite{costajussa:2017}.
%Finally, it is worth to mention that there is the Chispa Android application and web service\footnote{http://www.chispa.me}, that can be useful to tourists or traveling between countries, which use these languages \cite{centelles-2014}.

\section{Machine Translation Architecture}
%MARTA

\label{sec:mt}

In this section, we review the baseline system which is a standard phrase-based MT system and explain the architecture that we are using by dividing the problem of translation into: morphologically simplified translation and morphology generation.

\subsection{Phrase-based MT baseline system}

The popular phrase-based MT system \cite{koehn:2003} focuses on finding the most probable target text given a source text. In the last 20 years, the phrase-based system has dramatically evolved introducing new techniques and modifying the architecture; for example, replacing the noisy-channel for the log-linear model which combines a set of feature functions in the decoder, including the translation and language model, the reordering model and the lexical models. There is a widely used open-source software, \textit{Moses} \cite{moses}, which englobes a large community that helps in the progress of the system. As a consequence, phrase-based MT is a commoditized technology used at the academic and commercial level. However, there are still many challenges to solve, such as morphology generation.

\subsection{Divide and conquer MT architecture: simplified translation and morphology generation}

Morphology generation is not always achieved in the standard phrase-based system. This may be due to the fact that phrase-based MT uses a limited source context information to translate. Therefore, we are proposing to follow a similar strategy to previous works \cite{toutanova:2008,formiga:2013}, where authors do a first translation from source to a morphology-based simplified target and then, use the morphology generation module that transforms the simplified translation into the full form output.

\section{Morphological Generation Module}
%MARTA & CARLOS

\label{sec:mg}

In order to design the morphology generation module, we have to decide the morphology simplification we are applying to the translation. Since we are focusing on Chinese-to-Spanish task and based on \cite{costajussa:2015}, the simplification which achieves the best trade-off among highest translation gain and lowest complexity of morphological generation is the simplification in number and gender. Table \ref{tab:simplification} shows examples of this simplification. The main challenge of this task is that number and gender are generated from words where this inflection information (both number and gender) has been removed beforehand.

\begin{table*}[ht]
\begin{center} {
\scriptsize
\begin{tabular}{l|l}
\hline
Es$_{num}$ &  decidir[VMIP3\textbf{N}0] examinar[VMN0000] el[DA0M\textbf{N}0] c\
uesti\'on[NCF\textbf{N}000] en[SPS00] el[DA0M\textbf{N}0] per\'iodo[NCM\textbf{\
N}000] de[SPS00]\\
& sesi\'on[NCF\textbf{N}000] el[DA0M\textbf{N}0] tema[NCM\textbf{N}000] titular\
[AQ0M\textbf{N}0] ``[Fp] cuesti\'on[NCF\textbf{N}000] relativo[AQ0F\textbf{N}0]\
 a[SPS00] el[DA0M\textbf{N}0] \\
& derecho[NCM\textbf{N}000] humano[AQ0M\textbf{N}0] ``[Fp] .[Fp] \\
Es$_{gen}$ &  decidir[VMIP3S0] examinar[VMN0000] el[DA0\textbf{G}S0] cuesti\'on\
[NC\textbf{G}S000] en[SPS00] el[DA0\textbf{G}S0] per\'iodo[NC\textbf{G}S000] de\
[SPS00]\\
& sesi\'on[NC\textbf{G}S000] el[DA0\textbf{G}S0] tema[NC\textbf{G}S000] titular\
[AQ0\textbf{G}S0] ``[Fp] cuesti\'on[NC\textbf{G}S000] relativo[AQ0\textbf{G}S0]\
 a[SPS00] el[DA0\textbf{G}S0] \\
& derecho[NC\textbf{G}S000] humano[AQ0\textbf{G}S0] ``[Fp] .[Fp] \\
Es$_{numgen}$ &  decidir[VMIP3\textbf{N}0] examinar[VMN0000] el[DA0\textbf{GN}0\
] cuesti\'on[NC\textbf{GN}000] en[SPS00] el[DA0\textbf{GN}0] per\'iodo[NC\textbf{GN}000] de[SPS00]\\
& sesi\'on[NC\textbf{GN}000] el[DA0\textbf{GN}0] tema[NC\textbf{GN}000] titular\
[AQ0\textbf{GN}0] ``[Fp] cuesti\'on[NC\textbf{GN}000] relativo[AQ0\textbf{GN}00\
] a[SPS00] el[DA0\textbf{GN}0] \\
& derecho[NC\textbf{GN}000] humano[AQ0\textbf{GN}0] ``[Fp] .[Fp] \\\hline
Es &  Decide examinar la cuesti\'on en el per\'iodo de sesiones el tema titulad\
o `` Cuestiones relativas a los derechos humanos '' .\\ \hline
\end{tabular} }
\end{center}
\caption{Example of Spanish simplification into number, gender and both}.
\label{tab:simplification}
\end{table*}

With these results at hand, we propose an architecture of the morphology generation module, which is language independent and it is easily generalizable to other simplification schemes.                                                       

\begin{figure}[h]                                                           
\begin{center}                                                                 
\includegraphics[width=0.4\textwidth]{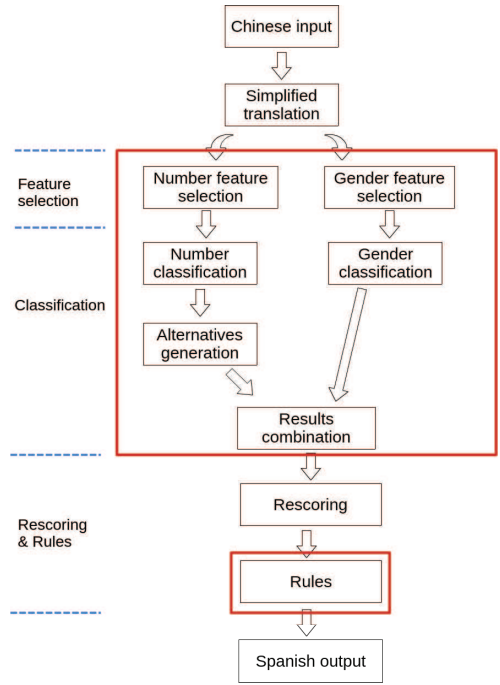}  
\end{center}                                                                   
\caption{\label{fig:procedure}Block Diagram for Morphology Generation}
\end{figure}                                                                   

The morphology generation procedure is summarised as follows and further detailed in the next subsections.

\begin{itemize}
\item Feature selection. We have investigated different set of features including information from both source and target languages. 

%As we will see in Section \ref{sec:evluation} best results are obtained when using a fix window of words from just the target language. From this, we are using only the word context of the sentence to predict the results. The use of only target features allows us to avoid errors produced by the word alignment if using source features. 

\item Classification. We propose a new deep learning classification architecture composed of different layers.%have tested different types of classification algorithms including neural networks and support vector machines. 

\item Rescoring and rules. We generate different alternatives of the classification output and rerank them using a language model. After, we use hand-crafted rules that allow to solve some specific problems. 

\end{itemize}

This procedure is depicted in Figure \ref{fig:procedure}, in which we can see that each of the above processes generates the needed input for the next step. Figure also shows in red the main subprocesses that have been developed on this work.% while the translation and rescoring tools where generated using \textit{Moses}. 

%\section{Classification Methodology}

\subsection{Feature selection}

We propose to compare several features for recovering morphological information. Given that both Chinese and simplified Spanish languages do not contain explicit morphology information, we start by simply using windows of words as source of information. We follow Collobert's approach in which each word is represented by a fixed size window of words in which the central element is the one to classify \cite{collobert2011natural}. 

In our case, we experiment with three different inputs: (1) Chinese window; (2) simplified Spanish window; (3) Spanish window adding information about its correspondant word in Chinese, i.e. information about pronouns and the number of characters in the word. The main advantage of the second one is that it is not dependant on the alignment file generated during translation.

%TO USE LATER 
Our classifiers did not have to train all types of words. Some types of words, such as prepositions (\textit{a, ante, cabo, de...}), do not have gender or number. Therefore our system was trained using only determiners, adjectives, verbs, pronouns and nouns which are the ones that present morphology variations in gender or number. However, note that all types of words are used in the windows. % complement and linkers between other words.

%For us, this structure has the following advantages: (1) having the word to classify in the center helps  the word is at the begining or the end of the sentence; (2) the fact of being fixed size windows provides as with an additional parameter to tune in the classifier. 
%\red{Finally we generate the vocabulary of our corpus and represent each word as its index in this vocabulary ordered by appearence. The size of this vocabulary is used as a parameter for our classifiers in order to trains the unkown word case. **ESTO TAMBIÉN EN LA SECCION 5**} 

%This process is splitted in two different steps. To generate the targets for the supervised training of the net we separately treat the gender and number windows based on the tag format used by \emph{freeling}.

\subsection{Classification architecture}

%Inspired by previous works \cite{collobert2011natural},  as features for our classifier we use windows of words as input. Each word is represented by a fixed size window of words in which the central element is the one to classify. We only use morphology-simplified target words to avoid dependence on alignment quality. 

\begin{figure}[h]                                                           
\begin{center}                                                                 
\includegraphics[width=0.4\textwidth]{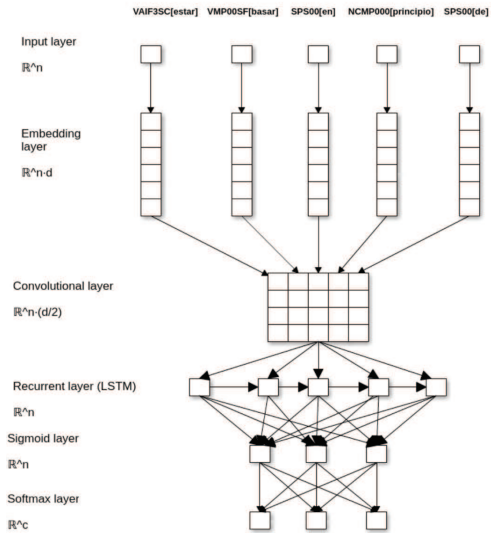}
\end{center}                                                                   
\caption{Neural network overview.}                          
\label{fig:architecture}                                                     
\end{figure}

\paragraph{Description} We propose to train two different models: one to retrieve gender and another to retrieve number. Each model decides among three different classes. Classes for gender classifier are masculine ($M$), femenine ($F$) and none ($N$);
and classes for number classifier are singular ($S$), plural ($P$) and none ($N$) \footnote{\textit{None} means that there is no need to specify gender or number information because the word is invariant in these terms. This happens for types of words (determiners, nouns, verbs, pronouns and adjectives) that can have gender and number in other cases.}. Again, we inspire our architecture in previous Collobert's proposal and we modify it by adding a recurrent neural network. This recurrent neural network is relevant because it keeps information about previous elements in a sequence and, in our classification problem, context words are very relevant. As a recurrent neural network, we use a Long Short Term Memory (LSTM) \cite{hochreiter1997long} that is proven efficient to deal with sequence NLP challenges \cite{sutskever:2014}. %that keeps information about the previous elements in the sequence. 
This kind of recurrent neural network is able to maintain information for several elements in the sequence and to forget it when needed. 
Figure \ref{fig:architecture} shows an overview of the different layers involved in the final classification architecture, which are detailed as follows:

\textit{Embedding.} We represent each word as its index in the vocabulary, i.e. every word is represented as one discrete value: 

$E(w) = W, W\in\mathbb{R}^{d}$, 

$w$ being the index of the word in the sorted vocabulary and $d$ the user chosen size of the array. Then, each word is represented as a numeric array and each window is a matrix.

\textit{Convolutional.} We add a convolutional neural network. This step allows the system to detect some common patterns between the different words. This layer's input consists in $W^{l}$ matrix of multidimensional arrays of size $n\cdot d$, where $n$ is the window length (in words) and $d$ is the size of the array created by the previous embedding layer. This layer's output is a matrix of the same size as the input.

\textit{Max Pooling.} This layer allows to extract most relevant features from the input data and reduces feature vectors to half.   

\textit{LSTM.} Each feature array is treated individually, generating a fixed size representation $h_{i}$ of the $i$th word using information of all the previous words (in the sequence). This layer's output, $h$, is the result of the last element of the sequence using information from all previous words. 
%The importance of this layer is its ability to maintain information of a sequence but also its ability to forget when the information stored is no longer useful. An example of this is the case of punctuation marks. When a punctuation mark is encountered in the text, the following words are not depending on the words before. 

\textit{Sigmoid.} This layer smoothes results obtained by previous layer and compresses results to the interval $[-1,1]$. This layer's input is a fixed size vector of shape $1 \cdot n$ where $n$ is the number of neurons in the previous LSTM layer. This layer's output is a vector of length $c$ equal to the number of classes to predict.

\textit{Softmax.} This layer allows to show results as probabilities by ensuring that the returned value of each class belongs to the $[0,1)$ interval and all classes add up 1.

\paragraph{Motivation} Our input data is PoS tagged and morphogically simplified before the classification architecture which largely reduces the information that can be extracted from individual words in the vocabulary. In addition, we can encounter out-of-vocabulary words for which no morphological information can be extracted.

The main source of information available is the context in which the word can be found in the sentence. Considering the window as a sequence enforces the behaviour a human would have while reading the sentence. The information of a word consists in itself and the words that surround it. Sometimes information preceeds the word and sometimes information is after the word. Words (like adjetives), which are modifying or complementing another word, generally take information from preceeding words. For example, in the sequence \textit{casa blanca}, the word \textit{blanca} could also be \textit{blanco, blancos or blancas} but because noun and adjective are required to have gender and number agreement, the femenine word \textit{casa} forces the femenine for \textit{blanca}. While, for example, determiners usually take information from posterior words. This fact motivates that the word to classify has to be placed at the center of the window.

Finally, given that we rely only on the context information since words themselves may not have any information, makes the recurrent neural network a key element in our architecture. The output $h$ of the layer can be considered a context vector of the whole window maintaining information of all the previously encountered words (in the same window).

\subsection{Rescoring and rules}

At this point in the pipeline, we have two models (gender and number) that allow us to generate the full Part-of-Speech (PoS) tag by combining the best results of both classifiers.

However, in order to improve the overall performance, we add a rescoring step followed by some hand-crafted rules. To generate the different alternatives, we represent every sentence as a graph (see Figure \ref{fig:example}), with the following properties:

\begin{figure}[h]                                                           
\begin{center}                                                                 
\includegraphics[width=0.2\textwidth]{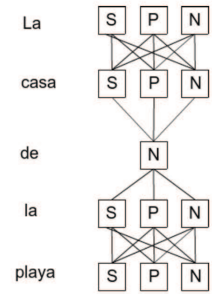}
\end{center}                                                                   
\caption{Example of sentence graph. S stands for singular, P for plural and N for none}
\label{fig:example}                                                     
\end{figure}     

\begin{itemize}
\item Each word is represented as a layer of the graph and each node represents a class of the classification model. 
\item A node only has edges with all the nodes of the next layer. This way we force the original sentence order. 
\item An edge's weight is the probability given by the classification model.
\item Each accepted path in the graph starts in the first layer and ends in the last one. This acyclic structure finds the best path in linear time, due to the fact that it goes through all layers and it picks the node with the greatest weight. One layer can have either 1 element (the word does not need to be classified, e.g. prepositions) or 3 elements (the word needs to be classified among the three number or gender categories).
\item Add the weight of a previously trained target language model.
\end{itemize}

We used Yen's algorithm \cite{yen1971finding} to find the best path, which has an associated cost of $O(KN^2logN)$, being $K$ the number of paths to find. 

See pseudo code in Algorithm \ref{pseudo}, where $A$ is the set paths chosen in the graph. The algorithm ends when this $A$ set contains $K$ paths or no further paths available to explore. $B$ contains all suboptimal paths that can be elected in future iterations.

There are two special cases that the models were not able to treat and we apply specific rules: (1) conjunctions \textit{y} and \textit{o} are replaced by \textit{e} and \textit{u} if they are placed in front of vowels. This could not be generated during translation because both words share the same tag and lemma; (2) verbs with a pronoun as a suffix, \textit{producirse}, second to last syllabe stretched (\textit{palabras llanas}) and ending in a vowel are not accentuated. However, after adding the suffix, these words should be accentuated because they become \textit{palabras esdrújulas}, which happen to be always accentuated.

\begin{algorithm}
\caption{Pseudo-code for k-best paths generation.}\label{pseudo}
\scriptsize
 \KwData{G Graph of the sentence, K}
 \KwResult{best k paths in  G }
 initialization\;
 A[0] = bestPath(G,0,final)\;
 B = [] \;
 i = 0\;
 \For{i \textless K}{
  \For{ i in range(0, len(A[K-1])-1)} {
      spurNode = A[K-1][i]\;
      root = A[K-1][0;i]\;
      \For {path in A} {
     	 \If {root = path[0:i]} {
   		 remove edge(i-1,i) from G\; 		 
     	 }    
      }
      \For {node in root and node != spurNode} {
   	 removes node node from  G\; 	 
      }
      spurPath = bestPath(G,spurnode, final)
      totalPath = root + spurPath\;
      B.append(totalPath)\;
      restore edges from G\;
      restore nodes from G;
      
      \If { B is empty}{
     	 break\;
      }
	B.sort()\;
      A.append(B[0])\;
      B.pop()\;    
  }
 }
\end{algorithm}

%CARLOS

\section{Experimental framework}
%MARTA

\label{sec:experimental}

In this section, we describe the data used for experimentation together with the corresponding preprocessing. In addition, we detail chosen parameters for the MT system and the classification algorithm.

\subsection{Data and preprocessing}

\begin{table}
    \centering
    \scriptsize
    \begin{tabular}{|l|l|l|l|l|l|}
    \hline
    L & \multicolumn{2}{l|}{Set} & S & W & V \\
    \hline
    \multirow{3}{1cm}{ES} & \multirow{2}{6mm}{Train} & Small & 58.6K & 2.3M & 22.5K \\ \cline{3-6}
    & & Large & 3.0M & 51.7M & 207.5K \\ \cline{2-6}
    & \multicolumn{2}{l|}{Development} & 990 & 43.4K & 5.4k \\ \cline{2-6}
    & \multicolumn{2}{l|}{Test} & 1K & 44.2K & 5.5K\\ \hline
%    &  & BTEC & 729 & 4.7K & 888\\ \hline \hline%\cline{3-6}                    
%    &  &Bible & 1K & 27.4K & 4.4K\\                                            
    \multirow{3}{1cm}{ZH} & \multirow{2}{6mm}{Train} & Small  & 58.6K & 1.6M & 17.8K \\ \cline{3-6}
    & & Large & 3.0M & 43.9M & 373.5K \\ \cline{2-6}
    & \multicolumn{2}{l|}{Development} & 990 & 33K & 3.7K \\ \cline{2-6}
    &\multicolumn{2}{l|}{Test} & 1K & 33.7K & 3.8K\\ \hline
%    & & BTEC & 729 & 4.1K & 737\\ \hline\hline %\cline{3-6}                     
%    & &Bible & 1K & 24.8K & 3.9K \\                                            
 %   \multirow{4}{1cm}{ZH Chars} & \multirow{2}{6mm}{Train} & Small & 58.6K & 2.8M & 3.8K \\ \cline{3-6}
 %   & & Large & 3.0M & 71.1M & 43.7K \\ \cline{2-6}
 %   & \multicolumn{2}{l|}{Development} & 990 & 53.9K & 1.7K \\ \cline{2-6}
 %   & \multirow{3}{6mm}{Test}& UN & 1K & 55.1K & 1.7K\\ \cline{3-6}
%    & & BTEC & 729 & 5.5K & 668\\ \hline
%    & &Bible & 1K & 34.5K & 1.8K\\ \hline                                      
    \end{tabular}
    \caption{Corpus Statistics. Number of sentences (S),words (W), vocabulary (V). M stands for millions and K stands for thousands.}
    \label{tabla:corpus}
\end{table}

One of the main contributions of this work is using the Chinese-Spanish language pair. 
In the last years, there has appeared more and more resources for this language pair available in \cite{unlarge} or from TAUS corporation\footnote{http://www.taus.net}. Therefore, differently from previous works on this language pair, we can test our approach in both a small and large data sets. 

\begin{itemize}
\item A small training corpus by using the United Nations Corpus (UN) \cite{un}.
\item A large training corpus by using, in addition to the UN corpus, the TAUS corpus, the Bible corpus \cite{chew:2006} and the BTEC (Basic Traveller Expressions Corpus) \cite{Takezawa:2006}. The TAUS corpus is around 2,890,000 sentences, the Bible corpus about 30,000 sentences and the BTEC corpus about 20,000 sentences.
\end{itemize}

Corpus statistics are shown in Table \ref{tabla:corpus}. %In the case of word segmentation, the size of the vocabulary is similar to the target vocabulary, while in the case of using Chinese characters, we have a much lower vocabulary. 
Development and test sets are taken from UN corpus. %We use the UN test set.

Corpus preprocessing consisted in tokenization, filtering empty sentences and longer than 50 words, Chinese segmentation by means of the ZhSeg \cite{zhseg}, Spanish lowercasing, filtering pairs of sentences with more than 10\% of non-Chinese characters in the Chinese side and more than 10\% of non-Spanish characters in the Spanish side.  Spanish PoS tagging was done using \textit{Freeling} \cite{padro12}. All chunking or name entity recognition was disabled to preserve the original number of words.% and the structure of the sentence. 

%It was also important break fixed phrases into individual words to ensure that words next to the get information to be processed.% NO entiendo...

\subsection{MT Baseline}

\textit{Moses} has been trained using default parameters, which  include:
grow-diag-final word  alignment  symmetrization, lexicalized reordering,  relative  frequencies  (conditional  and
posterior  probabilities)  with  phrase  discounting, lexical  weights,  phrase  bonus, accepting phrases up to  length  10, 
5-gram language model with kneser-ney smoothing, word bonus and MERT optimisation.

\subsection{Classification parameters}

To generate the classification architecture we used the library \textit{keras} \cite{chollet2015keras} for creating and ensambling the different layers. Using NVIDIA GTX Titan X GPUs with 12GB of memory and 3072 CUDA Cores, each classifier is trained on aproximately 1h and 12h for the small and large corpus, respectively.

Regarding classification parameters, experimentation has shown that number and gender classification tasks have different requirements. Table \ref{tabla:param} summarizes these parameters. 
The best window size is 9 and 7 words for number and gender, respectively. 
In both cases increasing this size lowers the accuracy of the system. 
The vocabulary size is fixed as a trade-off between giving enough information to the system to perform the classification while removing enough words to train the classifier for unknown words. 
The embedding size of 128 results in stable training, while further increasing this value augmented the training time and hardware cost. The filter size in the convolutional layer reached best results when it was slightly smaller than the window size, being 7 and 5 the best values for number and gender classification, respectively. Finally, increasing LSTM nodes up to 70 improved significantly for both classifiers.

\begin{table}[!h]
    \scriptsize
    \begin{center}
      \caption{Values of the different parameters of the classifiers}
        \begin{tabular}{|l|l|l|l|l|}
            \hline
            Parameter  & \multicolumn{2}{|c|}{Small} & \multicolumn{2}{|c|}{Large}  \\ \hline
            & Num & Gen & Num & Gen  \\ \hline
            Window size & 9 & 7   & 9 & 7 \\ \cline{1-5}
			Vocabulary size & 7000 & 9000   & 15000 & 15000 \\ \cline{1-5}
			Embedding  & 128 & 128 & 128  & 128 \\ \cline{1-5}
			Filter size & 7 & 5 & 7 & 5 \\ \cline{1-5}
			LSTM nodes  & 70 & 70   & 70 & 70 \\ \hline
         \end{tabular}
        \label{tabla:param}
    \end{center}
\end{table}

For windows, we only used the simplified Spanish translation. In Table \ref{my-label} we can see that testing different sources of information with the classifier of number for the small corpus. Adding Chinese has a negative effect in the classifier accuracy.

\begin{table}[]
\centering
\scriptsize
\caption{\label{my-label}Accuracy of the classifier of number using different sources of information.}
\begin{tabular}{|l|l|}
\hline
Features                 & Accuracy (\%) \\ \hline
Chinese window           & 72            \\ \hline
Spanish window           & 93,4          \\ \hline
Chinese + Spanish window & 86 \\ \hline            
\end{tabular}
\end{table}

\subsection{Rescoring and Full form generation}

As a rescoring tool, we use the one available in \textit{Moses} \footnote{https://github.com/moses-smt/mosesdecoder/tree/master/scripts/nbest-rescore}. We trained a standard n-gram language model with the SRILM toolkit \cite{stolcke:2002}. 

In order to generate the final full form we use the full PoS tag, generated from the postprocessing step, and the lemma, taken from the morphology-simplified translation output. Then, we use the vocabulary and conjugations rules provided by \textit{Freeling}.
\textit{Freeling}'s coverage raises the 99\%. When a word is not found in the dictionary, we test all gender and/or number inflections in descendant order of probability until a match is found. If none matched, the lemma is used as translation, which usually happens only in the case of cities or demonyms.

\section{Evaluation}
%CARLOS

\label{sec:evaluation}

In this section we discuss the results obtained both in classification and in the final translation task. 

Table \ref{tabla:resultadoscla} shows results for the classification task both number and gender and with the different corpus sets. We have contrasted our proposed classification architecture based on neural networks with standard machine learning techniques such as linear, cuadratic and sigmoid kernels SVMs \cite{cortes1995support}, random forests \cite{breiman2001random}, convolutional\cite{fukushima1980} and LSTM\cite{hochreiter1997long} neural networks (NN). All algorithms were tested using features and parameters described in previous sections with the exception of random forests in which we added the one hot encoding representation of the words to the features.
%   that of all the techniques we have tested the architecture consisting in combining convolutional and recurrent neural networks outperforms the rest of techniques tested.

We observe that our proposed architecture achieves by large the best results in all tasks. It is also remarkable that the accuracy is lower using the bigger corpus, this is due to the fact that the small set consisted in texts of the same domain and the vocabulary had a better representation of specific words such as country names.% that does not follow a rule to determine their morfology. 

\begin{table}[!h]
    \scriptsize
    \begin{center}
      \caption{Classification results. In bold, best results. Num stands for Number and Gen, for Gender}
        \begin{tabular}{|l|l|l|l|l|}
            \hline
            Algorithm  & \multicolumn{2}{|c|}{Small} & \multicolumn{2}{|c|}{Large}  \\ \hline
            & Num & Gen & Num & Gen  \\ \hline
            Naive Bayes & 61.3 & 53.5   & 61.3 & 53.5 \\ \cline{1-5}
			Lineal kernel SVM   & 68.1 & 71.7   & 65.8 & 69.3 \\ \cline{1-5}
			Cuadratic kernel SVM  & 77.8 & 81.3 & 77.6  & 82.7 \\ \cline{1-5}
			Sigmoid kernel SVM  & 83,1 & 87.4   & 81.5 & 84.2 \\ \cline{1-5}
			Random Forest  & 81.6 & 91.8   & 77.8 &  88.1 \\ \cline{1-5}
			Convolutional NN  & 81.3 & 93.9 & 83.9 & 94.2 \\ \cline{1-5}
			LSTM NN  & 68.1 & 73.3 & 70.8 & 71.4 \\ \cline{1-5}
			CNN + LSTM   & \textbf{93.7} & \textbf{98.4} & \textbf{88.9} & \textbf{96.1} \\ \hline
         \end{tabular}
        \label{tabla:resultadoscla}
    \end{center}
\end{table}

Tabla \ref{tabla:resultados} shows translation results. We show both the Oracle and the result in terms of METEOR \cite{banerjee2005meteor}. We observe improvement in most cases (when classifying number, gender, both and rescoring), but best results are obtained when classifying number and gender and rescoring number in the large corpus, obtaining a gain up to +0.7 METEOR.

%About the generation results we can see that are fairly superior in the large corpus. 

%Large corpus offers a bigger range of gain (see oracle), which allows 
%The main reason for this difference is the better results obtained by the simplified translation using a larger corpus. We also can see that the difference between the baseline and the oracle translation if bigger giving more margin for improvement to the classifiers.

%Note that the addition of the rescoring step improves results when applied to number but decreases when both characteristics are combined. This could be caused by the fact that the gender results are better from the classifier output. Also, rescoring may be difficult when balancing two non-normalised features. 

Rescoring step improves final results. Note that rescoring was only applied to number classification because gender classification model has a low classification error (bellow 2\%) which makes it harder to further decrease it. Additionally, gender and number classification scores are not be comparable and not easily integrated in Yen's algorithm.

\begin{table}[!]
    \scriptsize
    \begin{center}
      \caption{METEOR results. In bold, best results. Num stands for Number and Gen, for Gender}
        \begin{tabular}{|l|l|l|l|}
            \hline
            Set & System & \multicolumn{2}{|c|}{UN} \\ \hline
& & Oracle & Result \\ \hline
Small & Baseline & - & 55.29    \\ \cline{2-4}
      & +Num  & 55.60 & 55.35    \\ \cline{2-4}
& +Gen  & 55.45 & 55.39    \\ \cline{2-4}
& +Num\&Gen  & 56.81 & 55.48    \\ \cline{2-4}
& +Num\&Gen +Rescoring(Num\&Gen)  & - & 54.91    \\ \cline{2-4}
& +Num\&Gen +Rescoring(Num)   & - & \textbf{55.56}    \\ \hline
Large & Baseline & - &  56.98   \\ \cline{2-4}
      & +Num  &  58.87 &  57.51   \\ \cline{2-4}
& +Gen  & 57.56  &  57.32   \\ \cline{2-4}
& +Num\&Gen  & 62.41 &  57.13  \\ \cline{2-4}
%& +Num\&Gen +Rescoring(Num\&Gen)  & - &  56.47   \\ \cline{2-4}
& +Num\&Gen Rescoring   & - &  \textbf{57.74}   \\ \hline
         \end{tabular}
        \label{tabla:resultados}
    \end{center}
\end{table}

\section{Conclusions}
%MARTA

\label{sec:conclusions}

Chinese-to-Spanish translation task is challenging, specially because of Spanish being morphologically rich compared to Chinese.  Main contributions of this paper include correctly de-coupling the translation and morphological generation tasks and proposing a new classification architecture, based on deep learning, for number and gender. 

Standard phrase-based MT procedure is changed to first translating into a morphologically simplified target (in terms of number and gender); then, introducing the classification algorithm, based on a new proposed neural network-based architecture, that retrieves the simplified morphology; and composing the final full form by using the standard \textit{Freeling} dictionary. 

Results of the proposed neural-network architecture in the classification task compared to standard algorithms (SVM or random forests) are significantly better and results in the translation task achieve up to 0.7 METEOR improvement. As further work, we intend to further simplify morphology and extend the scope of the classification.

\section*{Acknowledgements}

This work is supported by the 7th Framework Program of the European Commission through the International Outgoing Fellowship Marie Curie Action (IMTraP-2011-29951) and also by the Spanish Ministerio de Econom\'ia y Competitividad and Fondo Europeo de Desarrollo Regional, through contract TEC2015-69266-P (MINECO/FEDER, UE) and the postdoctoral senior grant \textit{Ram\'on y Cajal}.

%\newpage

\bibliography{eacl-paper}
\bibliographystyle{acl2016}

\end{document}